\documentclass[acmtog, authorversion,nonacm]{acmart}

\usepackage{graphicx} \graphicspath{{figures/}}
\usepackage[linesnumbered,ruled,vlined]{algorithm2e} 

\SetAlFnt{\small}
\SetAlCapFnt{\small}
\SetAlCapNameFnt{\small}
\SetAlCapHSkip{0pt}
\usepackage{acronym}
\usepackage{enumitem}
\usepackage{balance}
\usepackage{xspace}
\usepackage{setspace}
\usepackage[skip=3pt,font=small]{subcaption}
\usepackage[skip=3pt,font=small]{caption}
\usepackage[capitalise,noabbrev,nameinlink]{cleveref}
\usepackage{booktabs,tabularx,colortbl,multirow,multicol,array,makecell,tabularray}
\usepackage{overpic,wrapfig}
\usepackage[misc]{ifsym}
\usepackage{pifont}

\makeatletter
\DeclareRobustCommand\onedot{\futurelet\@let@token\@onedot}
\def\@onedot{\ifx\@let@token.\else.\null\fi\xspace}
\def\eg{\emph{e.g}\onedot} 

\def\ie{\emph{i.e}\onedot}

\makeatother

\acrodef{hsi}[HSI]{Human-Scene Interaction}
\acrodef{ik}[IK]{Inverse Kinematic}
\acrodef{hoi}[HOI]{Human-Object Interaction}
\acrodef{cvae}[cVAE]{conditional Variational Auto-Encoder}
\acrodef{icp}[ICP]{Iterative Closest Points}
\acrodef{mocap}[MoCap]{motion-captured}
\acrodef{vit}[ViT]{Vision Transformer}
\acrodef{fid}[FID]{Fréchet Inception Distance}

\newcommand{\model}{\textbf{\mbox{DreamArt}}\xspace}
\begin{document}
\title{Generating Interactable Articulated Objects from a Single Image}

\author{Ruijie Lu}
\affiliation{%
  \institution{State Key Lab of General AI, Peking University}
  \country{China}}
\email{jason_lu@pku.edu.cn}
\authornote{Work done as an intern at BIGAI}

\author{Yu Liu}
\affiliation{%
  \institution{Tsinghua University}
  \country{China}}
\email{liuyu_ai@foxmail.com}

\author{Jiaxiang Tang}
\affiliation{%
  \institution{State Key Lab of General AI, Peking University}
  \country{China}}
\email{tjx@pku.edu.cn}

\author{Junfeng Ni}
\affiliation{%
  \institution{Tsinghua University}
  \country{China}}
\email{njf23@mails.tsinghua.edu.cn}

\author{Yuxiang Wang}
\affiliation{%
  \institution{State Key Lab of General AI, Peking University}
  \country{China}}
\email{yuxiang123@stu.pku.edu.cn}

\author{Diwen Wan}
\affiliation{%
  \institution{State Key Lab of General AI, Peking University}
  \country{China}}
\email{wan@stu.pku.edu.cn}

\author{Gang Zeng}
\affiliation{%
  \institution{State Key Lab of General AI, Peking University}
  \country{China}}
\email{gang.zeng@pku.edu.cn}
\authornote{Corresponding authors}

\author{Yixin Chen}
\affiliation{%
  \institution{State Key Lab of General AI, BIGAI}
  \country{China}}
\email{ethanchen@g.ucla.edu}
\authornotemark[2]

\author{Siyuan Huang}
\affiliation{%
  \institution{State Key Lab of General AI, BIGAI}
  \country{China}}
\email{syhuang@bigai.ai}
\authornotemark[2]

\begin{abstract}
Generating articulated objects, such as laptops and microwaves, is a crucial yet challenging task with extensive applications in Embodied AI and AR/VR. Current image-to-3D methods primarily focus on surface geometry and texture, neglecting part decomposition and articulation modeling. Meanwhile, neural reconstruction approaches (\eg, NeRF or Gaussian Splatting) rely on dense multi-view or interaction data, limiting their scalability. 
In this paper, we introduce \model, a novel framework for generating high-fidelity, interactable articulated assets from single-view images. \model employs a three-stage pipeline: firstly, it reconstructs part‑segmented and complete 3D object meshes through a combination of image-to-3D generation, mask-prompted 3D segmentation, and part amodal completion. Second, we fine-tune a video diffusion model to capture part-level articulation priors, leveraging movable part masks as prompt and amodal images to mitigate ambiguities caused by occlusion.
Finally, \model optimizes the articulation motion, represented by a dual quaternion, and conducts global texture refinement and repainting to ensure coherent, high-quality textures across all parts.
Experimental results demonstrate that \model effectively generates high-quality articulated objects, possessing accurate part shape, high appearance fidelity, and plausible articulation, thereby providing a scalable solution for articulated asset generation. Our project page is available at \href{https://dream-art-0.github.io/DreamArt/}{https://dream-art-0.github.io/DreamArt/}.
\end{abstract}

%
%



%
%


\begin{teaserfigure}
  \includegraphics[width=\linewidth]{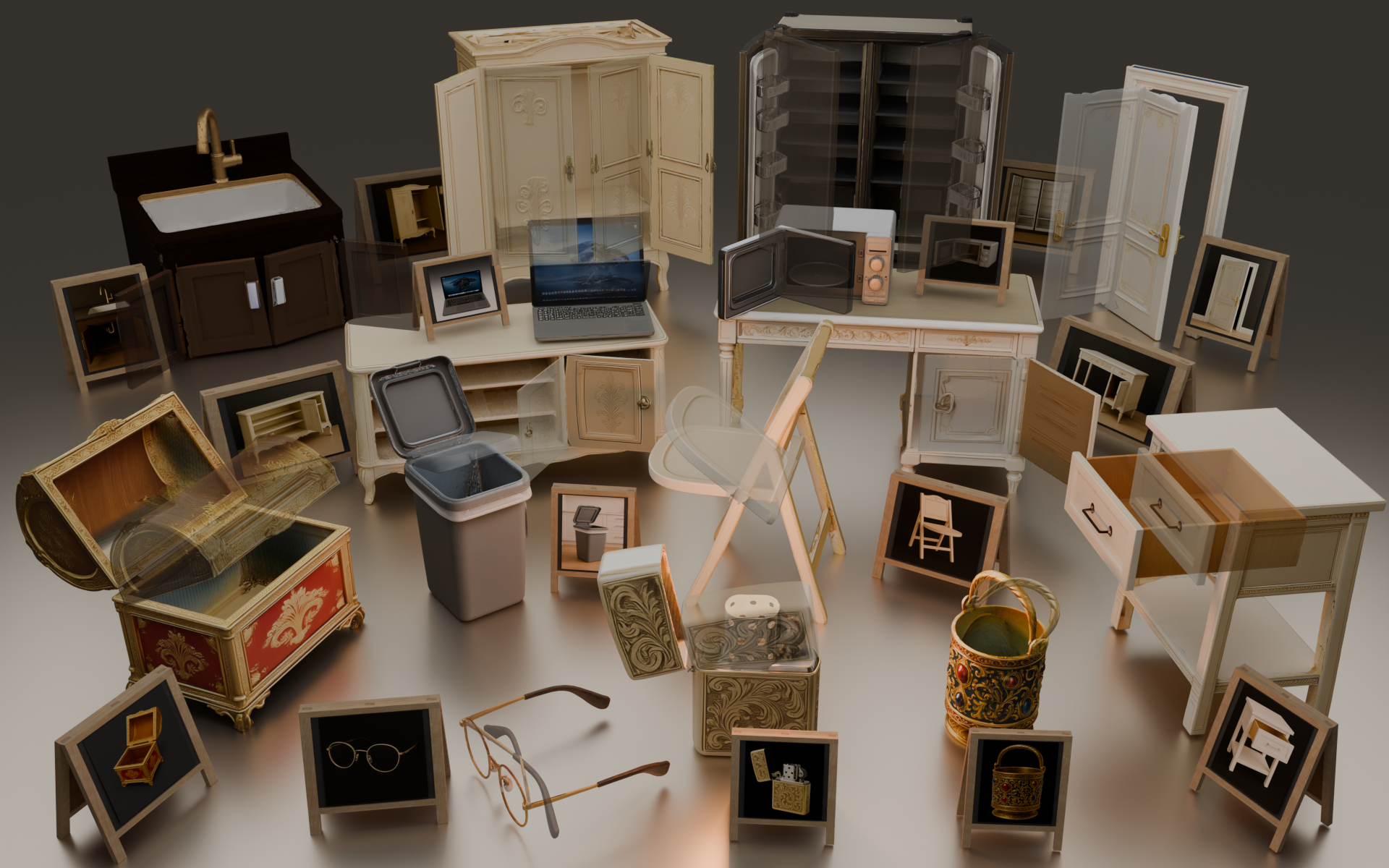}
  \caption{We propose \model, a framework for generating interactable articulated objects from a single-view image. Our method generalizes across diverse object categories, producing well-segmented parts and physically plausible articulations. All results are from in-the-wild images.}
  \label{fig:teaser}
\end{teaserfigure}

\maketitle

\section{Introduction}
Articulated objects (\eg, laptops, microwaves) are ubiquitous in everyday environments, making the generation of high-fidelity, interactable 3D assets at scale increasingly important across a wide range of applications, including embodied AI (EAI)~\cite{huang2023embodied, duan2022survey, brohan2022rt, huang2025unveiling, huang2025leo, jiang2024autonomous}, robotics~\cite{eisner2022flowbot3d, xiao2025robot, li2024ag2manip, luo2024physpart, li2024flowbothd}, and scene synthesis and reconstruction~\cite{wang2024roomtex, liu2024slotlifter, yang2024physcene, yu2025metascenes, ni2024phyrecon, chen2024single, wan2024superpoint, shen2025trace3d, ni2025decompositional}. 
However, the creation of high-quality, interactable assets still heavily relies on manual efforts by professional artists, despite recent advances in 3D object generation~\cite{xiang2024structured, ye2025hi3dgen, chen2024dora, zhao2025hunyuan3d, zhang2024clay, li2025triposg, he2025sparseflex}.
The core challenge lies in the intrinsic complexity of articulated objects: they require both precise part-level 3D modeling and accurate reasoning about their articulation patterns. 

In this paper, we tackle the task of Image-to-Articulated-Asset (I2A) generation: creating high-fidelity, interactable 3D assets
from a single image. Recent approaches attempt to reconstruct articulated objects from videos~\cite{song2024reacto, wan2024template, peng2025generalizable} or multi-view images~\cite{liu2023paris, weng2024neural, liu2025artgs, deng2024articulate, jiang2022ditto, wang2025self, lin2025splart, xia2025drawer, yu2025part} that capture multiple articulation states. While these methods can model detailed part-level motions, they require dense temporal and spatial observations, making them expensive to collect and limiting their scalability. Another line of research~\cite{li2024dragapart, li2024puppet, gao2025partrm, shi2024dragdiffusion} explores learning part-level motion priors via user-provided ``drag'' prompts. While promising, these methods rely on manually specified motion cues (\ie, starting and ending point of the ``drag'') and suffer from ambiguity in the drag signal to clearly identify the correct moving part, especially in complex, multi-part scenarios. These limitations hinder their practical use in fully articulated asset generation. 

To address these limitations, we propose \model, a novel framework for I2A generation. Our key objectives include accurately recovering part-aware 3D object structure, reasoning about plausible part-level articulation patterns with minimal human cues, and seamlessly distilling these learned priors into 3D meshes for articulation optimization. \model explicitly addresses these objectives through a three-stage pipeline. 

Firstly, to recover part-aware 3D objects, we build upon recent advance in image-to-3D models~\cite{zhao2025hunyuan3d, xiang2024structured} and further augment them with 3D part segmentation~\cite{yang2024sampart3d, liu2025partfield, tang2024segment} and amodal completion for both geometry~\cite{yang2025holopart} and texture~\cite{tang2024intex, lugmayr2022repaint}. 

To infer plausible articulation patterns, we fine-tune a conditional video diffusion model to predict videos demonstrating object articulation from a single image, drawing inspiration from recent advances in repurposing video generation models~\cite{blattmann2023stable, hong2022cogvideo, kong2024hunyuanvideo, wang2025wan, agarwal2025cosmos} for novel tasks~\cite{van2024generative, xing2024dynamicrafter, lu2025taco, hu2024depthcrafter, shao2024learning, Ma2025See3D}.  Our key innovations lie in two aspects. First, we adopt a movable part mask as the visual prompt. Unlike the ``drag'' prompt~\cite{li2024puppet}, the mask avoids human intervention, readily available through off-the-shelf 2D segmentation models~\cite{kirillov2023segment, ravi2024sam}. It also naturally fits into the video diffusion framework as an image-based condition, and mitigates the ``drag'' ambiguity by clearly specifying the moving part, especially in multi-part scenarios. Secondly, occlusions between different parts are common in articulated objects (\eg, cabinet doors blocking internal structures). This requires the model to infer both the articulation and occluded regions to produce reasonable video content. To address this, we incorporate amodal images of movable and base parts, derived from our part-aware 3D object generation, as additional conditions. These cues guide the diffusion model to focus on learning articulation patterns while avoiding synthesizing blurry or conflicting contents.

Once the articulation video is obtained, we perform part-aware, mesh-based optimization to estimate the underlying articulation. We adopt a unified dual quaternion representation that parameterizes both prismatic and revolute joints, and estimate a timestep-dependent transformation to capture the motion magnitude of the movable part across the video.
The movable part mesh is deformed according to the joint type, axis, and motion magnitude, and we render the base and movable parts through a \textit{Differentiable Soft Depth Blending}. The articulation parameters are optimized via the rendering loss. 
Finally, we apply global texture refinement and repainting~\cite{zhao2025hunyuan3d, huang2024mv, podell2023sdxl, feng2025romantex, rombach2022high} to ensure high-quality, consistent textures across all object parts.

Extensive experiments demonstrate that our method effectively generates part-aware 3D object meshes, predicts plausible articulation videos, and produces interactable articulated assets in a unified pipeline. It generalizes well to in-the-wild images, outperforming existing approaches in both articulation video synthesis and articulated object generation, highlighting its potential for scalable articulated asset creation.

To sum up, our main contributions are:
\begin{enumerate}[nolistsep,noitemsep,leftmargin=*]
    \item We propose \model, a novel part and motion-aware framework that generates high-quality interactable articulated object meshes from a single-view image.
    \item We present a novel articulation video synthesis method that leverages movable part masks and amodal images as prompts, eliminating the need for manually specified motion directions and effectively resolving ambiguities in multi-part scenarios.
    \item Experiments show the efficacy of our approach in generating high-quality, interactable articulated assets with well-segmented part geometry, coherent appearance, and plausible articulations.
\end{enumerate}

\section{Related Work}
\subsection{3D/4D Assets Creation}
Early explorations on 3D asset generation primarily followed two paradigms: i) optimization-based approaches~\cite{qian2023magic123, wang2023prolificdreamer, yi2024gaussiandreamer, long2024wonder3d, chen2023fantasia3d, tang2023dreamgaussian, poole2022dreamfusion, liu2023syncdreamer}, which distill 2D generative priors~\cite{rombach2022high, liu2023zero, shi2023zero123++, shi2023mvdream, wang2023imagedream, lu2025movis} into 3D representations~\cite{mildenhall2021nerf, kerbl20233d}, and ii) feed-forward methods~\cite{hong2023lrm, tang2024lgm, xu2024grm, wu2024unique3d, liu2023one, liu2024one}.
More recently, latent diffusion~\cite{liu2022flow, lipman2022flow, ho2020denoising} in 3D space, often represented by features such as 3DShape2VecSet~\cite{zhang20233dshape2vecset}, has emerged~\cite{zhang2024clay, zhao2025hunyuan3d, xiang2024structured, chen2024dora, li2025triposg, he2025sparseflex} by incorporating large-scale training data from Objaverse~\cite{deitke2023objaverse_xl, deitke2023objaverse}. These models significantly improve geometry recovery quality; while they lack part structures and articulation modeling, they still offer a valuable foundation for building interactable articulated assets. 
Parallel to static 3D object generation, recent work explores text/image-to-4D generation~\cite{bahmani20244d, ling2024align, bahmani2024tc4d, li2024vivid, zhao2023animate124, ren2023dreamgaussian4d, liang2024diffusion4d, zhang20244diffusion}, which typically adopts off-the-shelf video generation models~\cite{blattmann2023stable, wang2023modelscope, zhang2023i2vgen} for score distillation, or directly estimate 4D representation from video inputs~\cite{xie2024sv4d, zeng2024stag4d, li2024dreammesh4d, wu2024sc4d, ren2024l4gm, yao2025sv4d}. However, existing video generation models often fail to produce temporally stable and physically plausible articulation motions, and the 4D optimization process lacks explicit articulation modeling, making it difficult to recover accurate articulation patterns. In contrast, our method incorporates explicit articulation modeling in both video generation and optimization to ensure physical plausibility.

\begin{figure*}[ht!]
  \includegraphics[width=\textwidth]{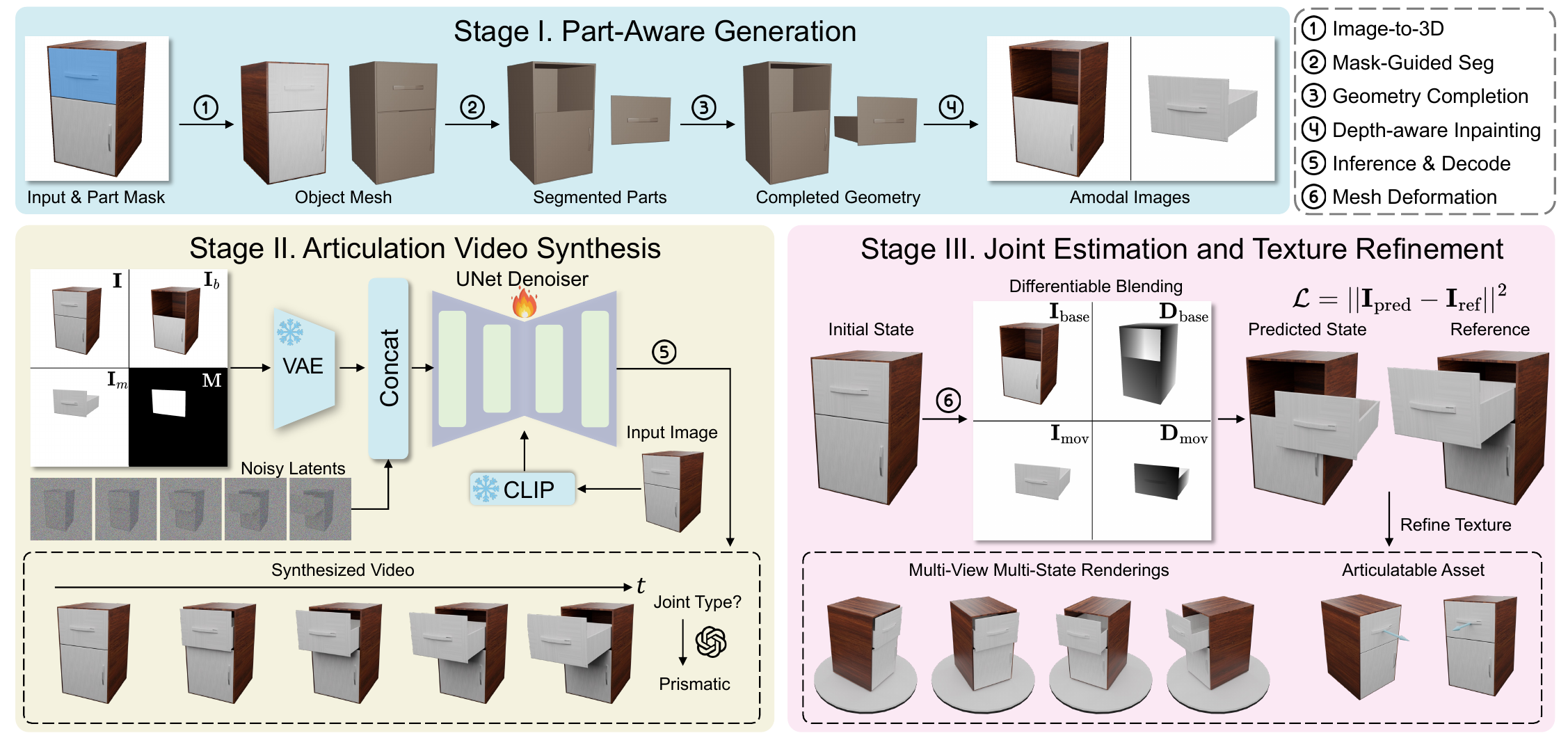}
  \caption{\textbf{Method Overview.} Our three-stage pipeline first reconstructs complete, segmented part meshes from a single image. Next, it synthesizes plausible articulation videos using amodal images and part masks as prompts. Finally, it optimizes joint parameters and refines texture maps for enhanced realism.}
  \label{fig:method}
\end{figure*}

\subsection{Articulated Object Modeling}
One line of research focuses on accurately reconstructing articulated objects (Digital Twins) from two-state multi-view images~\cite{liu2025artgs, weng2024neural, liu2023paris, wu2025reartgs, guo2025articulatedgs} or videos~\cite{song2024reacto, tu2025dreamo}. Although these methods are designed to capture articulation motion patterns precisely, they typically require dense observations of the same object, both temporally (multi-states) and spatially (multi-view images), limiting their scalability and applicability.
Another line of work explores procedural articulated object generation~\cite{chen2024urdformer, mandi2024real2code, liu2024cage, le2024articulate, liu2024singapo, qiu2025articulate, gao2024meshart}, via code generation, part mesh retrieval with vision-language models (VLMs), or hand-crafted rules~\cite{lian2025infinite}. However, these approaches face several limitations. Firstly, they often depend on predefined content such as part mesh libraries, code templates, VLM-predicted part connectivity graphs and joints, and heuristic rules, lacking generalization capability to diverse object categories. Secondly, they directly predict articulation parameters without dedicated refinement mechanisms, which leads to suboptimal accuracy.
In contrast, our method leverages synthesized articulation videos to optimize joint parameters through differentiable rendering at the mesh level, ensuring generalizability to in-the-wild cases with higher accuracy and more physically plausible articulation. 

\subsection{Part-level Articulation Prior}
Recent advances in image-to-video generation~\cite{blattmann2023stable, blattmann2023align, ge2023preserve, xing2024dynamicrafter, hong2022cogvideo, guo2023animatediff, agarwal2025cosmos, zhang2025packing, kong2024hunyuanvideo, wang2025wan, seawead2025seaweed} have significantly improved the quality and diversity of synthesized videos, naturally providing generic motion priors for articulated object generation. However, these models offer limited control, making it difficult to guarantee the physical plausibility of the synthesized articulation video.
To improve controllability, a line of research introduces ``drag-based'' prompts to guide part-level motion synthesis through images~\cite{li2024dragapart, shi2024dragdiffusion, pan2023drag, mou2023dragondiffusion} or videos~\cite{gao2025partrm, li2024puppet}. While intuitive, ``drag'' prompts present several challenges. Firstly, they rely on human-specified motion cues. Secondly, the ``drag'' prompts often exhibit ambiguity, as they may not clearly indicate motion type (\eg, prismatic vs. revolute) or identify the intended moving part, particularly in complex multi-part scenarios. 
More recently, ATOP~\cite{vora2025articulate} proposes learning finer part-level motion priors through independent video generation models for each object category. However, the category-specific design struggles when adapting to unseen object categories. In contrast, our model learns a more generalizable motion prior by leveraging a more intuitive and readily available control signal, the movable part mask.

\section{Methods}
In this section, we present \model, which aims to generate interactable articulated assets from a single-view image. The framework consists of three stages:
(1) \textbf{Part-Aware 3D Object Generation}, where we combine image-to-3D reconstruction with mask-guided 3D part segmentation and amodal completion techniques to produce a unified object mesh with complete and explicitly segmented parts (\cref{method:3d-gen});
(2) \textbf{Articulation Video Synthesis}, where we fine-tune a video diffusion model to synthesize plausible articulation videos conditioned on the single-view image, movable part mask, and corresponding amodal images (\cref{method:video-gen});
(3) \textbf{Joint Estimation and Texture Refinement}, where we optimize dual quaternion–based articulation parameters and refine textures to ensure both physically coherent motion and high-fidelity appearance (\cref{method:optimize}).
An overview is provided in \cref{fig:method}.

\subsection{Part-Aware 3D Object Generation}
\label{method:3d-gen}
Given a single-view image, we can readily obtain a high-fidelity, holistic textured mesh using advanced image-to-3D techniques~\cite{zhao2025hunyuan3d}, along with a 2D segmentation mask of the movable part derived from SAM~\cite{kirillov2023segment}. Let $\mathbf{I}$ denote the input image, $\mathbf{M}$ the 2D movable part mask, and $\mathbf{v}$ the reconstructed textured mesh. Our objective in this step is to derive the following: the complete movable part mesh $\mathbf{v}_m$, the complete base part mesh $\mathbf{v}_b$, and the amodal images of the movable part $\mathbf{I}_m$ and base part $\mathbf{I}_b$, both rendered from the same viewpoint as the input image.

\paragraph{Mask-Guided 3D Segmentation.} A well-segmented mesh is essential for articulated object modeling, as articulation is inherently defined at the part level. Existing 3D segmentation methods typically lift 2D features~\cite{oquab2023dinov2} into a volumetric field~\cite{yang2024sampart3d}, or perform clustering directly on a predicted 3D feature field with a specified number of parts~\cite{liu2025partfield}. However, both approaches struggle to precisely segment the target part, largely due to inherent ambiguities in part granularity. Alternatively, given a single-view mask, its corresponding faces can be identified by back-projection through differentiable mesh rendering~\cite{laine2020modular}. Yet, this strategy suffers from occlusion, as the target part mesh may not be visible from the input view, making it difficult to obtain a complete segmentation. 

Therefore, we adopt a hybrid approach that guides the clustering of mesh faces using the single-view mask image. Given the predicted $d$-dimensional per-face feature $\mathbf{F}_i \in \mathbb{R}^d$ from PartField~\cite{liu2025partfield} for each face $i$, and the set of faces $\mathcal{S}$ whose projections fall within the 2D movable part mask $\mathbf{M}$, we first compute the mean feature $\mathbf{F}_m$ of $\mathcal{S}$ as $\mathbf{F}_m = \frac{1}{|\mathcal{S}|} \sum_{i \in \mathcal{S}} \mathbf{F}_i$. Next, all mesh faces are classified into two groups by comparing their feature distance to $\mathbf{F}_m$. Specifically, face $i$ is assigned to the movable part if
\begin{equation}
||\mathbf{F}_i - \mathbf{F}_m||^2 \leq \max \limits_ {j \in \mathcal{S}} ||\mathbf{F}_j - \mathbf{F}_m||^2.
\label{eq:thresholding}
\end{equation}
Then, to remove the outlier and improve spatial smoothness, we apply k-means clustering to refine the segmentation using each part's feature centroids as initialization. This yields the segmented movable part mesh $\mathbf{v}_m^{\mathrm{partial}}$ and base part mesh $\mathbf{v}_b^{\mathrm{partial}}$ from the reconstructed textured mesh $\mathbf{v}$.

\paragraph{Part Amodal Completion.} The segmented part meshes are geometrically incomplete, as existing image-to-3D methods primarily focus on generating surface while ignoring internal and occluded structures (\eg, the drawer of the table is only partially observed in \cref{fig:method}). We first utilize HoloPart~\cite{yang2025holopart} to get the complete part mesh $\mathbf{v}_m, \mathbf{v}_b$ from the partial mesh $\mathbf{v}_m^{\mathrm{partial}}, \mathbf{v}_b^{\mathrm{partial}}$. Based on the complete mesh, we generate amodal images that fill in missing appearance information (\eg, the interior area of the drawer) on the input view using inpainting models $\mathcal{F}_{\text{inp}}$~\cite{tang2024intex, zhang2023adding}. Specifically, for the movable part: 
\begin{equation}
\mathbf{I}_{m} = \mathcal{F}_{\text{inp}}(\mathbf{I} \odot \mathbf{M}, \mathbf{M}_{\text{inp}}),
\label{eq:inpainting}
\end{equation}
where $\mathbf{I} \odot \mathbf{M}$ extracts the visible area for the movable part with pixel-wise production $\odot$, and $\mathbf{M}_{\text{inp}}$ stands for the area to be inpainted. The same process is applied to the base part as well.

\begin{figure*}[t!]
  \includegraphics[width=\linewidth]{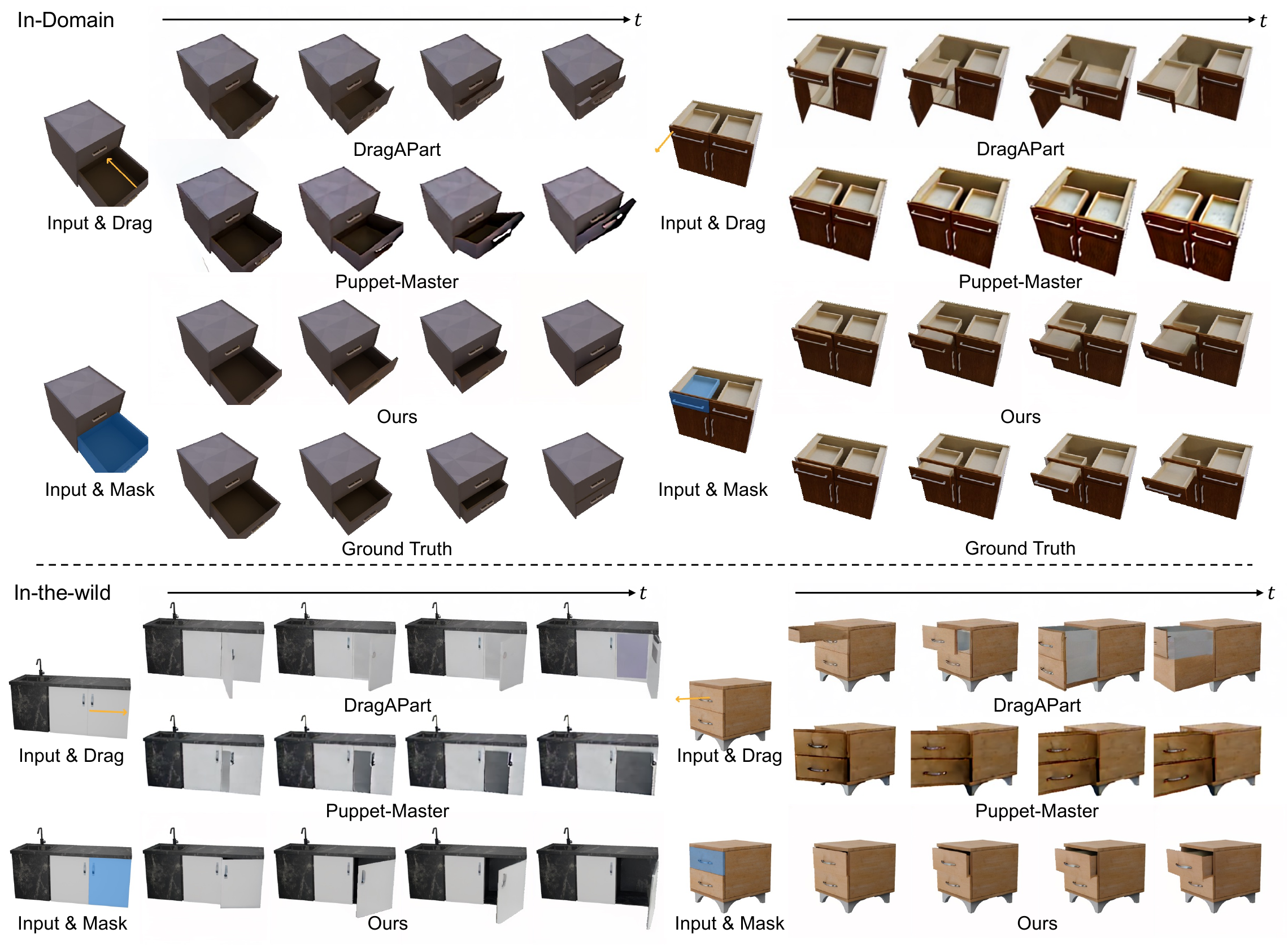}
  \caption{\textbf{Qualitative comparison of synthesized articulation videos.} We present qualitative results on both in-domain and in-the-wild data. Our method consistently outperforms the baselines by producing clearer and more plausible articulation, particularly in multi-part object scenarios.}
  \label{fig:quali}
\end{figure*}

\subsection{Articulation Video Synthesis}
\label{method:video-gen}
To infer plausible articulation patterns of an object, we propose to learn a generalizable video generator $f$ to synthesize articulation sequences, leveraging the rich spatio-temporal consistent manifolds of video diffusion models through Internet-scale video pretraining.

To steer the generation towards meaningful part-level articulations, we first introduce a task-specific prompt, the movable part mask $\mathbf{M}$, during fine-tuning. Compared to alternative forms of motion guidance, such as ``drag'' that rely on manually defined trajectories, the movable part mask offers several key advantages. First, it naturally fits into the video diffusion framework as an image-based condition, making it simple and efficient to integrate into the model architecture. Second, the mask prompt incorporates minimal human efforts while not affecting the reasoning of the underlying articulation patterns. Finally, it provides a less ambiguous signal for specifying the movable part, especially in multi-part scenarios.

However, learning articulation solely from RGB images poses inherent challenges due to frequent mutual occlusions between the base and movable parts. For example, a closed cabinet door initially occludes internal structures, which are gradually revealed as articulation unfolds. If such occluded content is entirely left to be hallucinated during video generation, the diffusion model may allocate extra capacity to shape and appearance synthesis while struggling to learn accurate articulation patterns. To mitigate this issue, we introduce amodal images $\mathbf{I}_m$ and $\mathbf{I}_b$ for both the movable and base part, as additional conditions to provide their fully visible information for the video generator $f$:
\begin{equation}
\mathbf{V} = f(\mathbf{I}, \mathbf{M}, \mathbf{I}_m, \mathbf{I}_b),
\label{eq:video-diffusion}
\end{equation}
where $\mathbf{V}$ stands for the articulation sequence. Specifically, all conditioning images are replicated $N$ times, where $N$ denotes the number of frames in the pre-trained video diffusion model~\cite{blattmann2023stable}. The replicated images are then encoded using the same VAE~\cite{kingma2013auto}, concatenated in the latent space, and fed into the denoising UNet~\cite{ronneberger2015u} together with the noisy articulation sequence $\mathbf{\hat{V}}$.

\begin{figure*}[ht]
  \includegraphics[width=\linewidth]{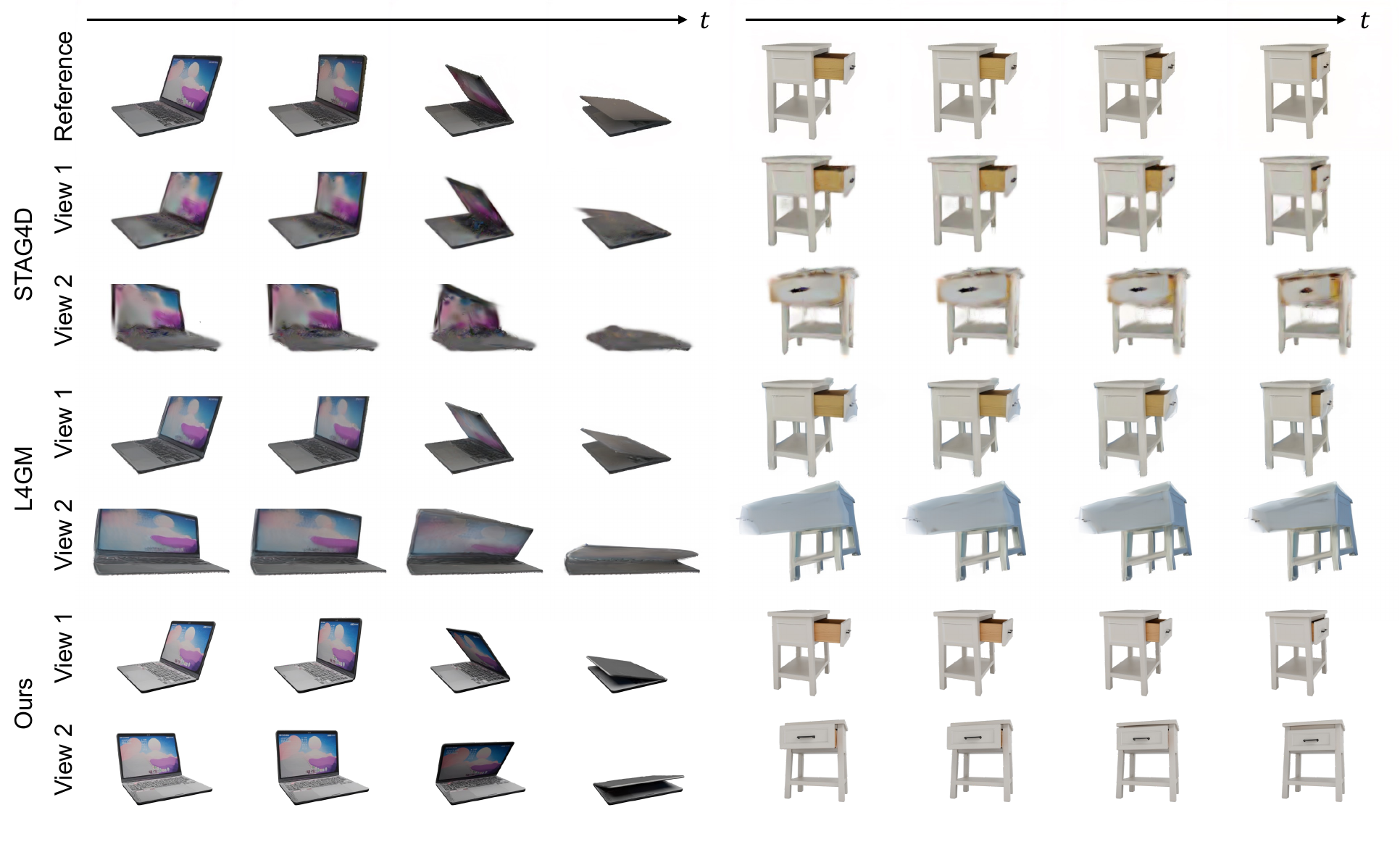}
  \caption{\textbf{Visualizations on asset synthesis.} \model shows clearer images with more plausible articulations than baselines, especially under novel views.}
  \label{fig:quali_comp}
\end{figure*}
\subsection{Joint Estimation and Texture Refinement}
\label{method:optimize}
Given the synthesized articulation video $\mathbf{V}$, our objective is to estimate the articulation parameters of the movable part. To this end, we first utilize a vision-language model (VLM) to infer the joint type (\ie, prismatic or revolute) from the video frames for articulation optimization. We represent the articulation axis with a joint position $\mathbf{A}_\text{pos} \in \mathbb{R}^3$ and an axis direction $\mathbf{A}_\text{dir} \in \mathbb{R}^3$. $\mathbf{A}_\text{pos}$ is not functionally required for prismatic joints, but is retained for formulation consistency.
We also leverage an MLP predictor $\mathcal{F}_{\text{motion}}(\cdot)$ to estimate the articulation motion offset $\theta_t$, which specifies the magnitude of translation or rotation of the movable part at timestep $t \in \{1, 2, \dots, N\}$ corresponding to the predicted articulation video:
\begin{equation}
\theta_t = \mathcal{F}_{\text{motion}}(t),
\label{eq:motion_mlp}
\end{equation}
The dual quaternion of the movable part, $\boldsymbol{q}_{r,t}$ and $\boldsymbol{q}_{d,t}$, represents the rotation and translation components, respectively:
\begin{equation}
\boldsymbol{q}_{d,t} = 
\begin{cases}
0.5 \cdot \mathbf{T} \otimes \boldsymbol{q}_{r,t}, & \text{prismatic} \\
0.5 \cdot (\mathbf{T} \otimes \boldsymbol{q}_{r,t} - \boldsymbol{q}_{r,t} \otimes \mathbf{T}), & \text{revolute}
\end{cases}
\label{dual-qua}
\end{equation}
where $\otimes$ denotes the quaternion product. For prismatic joints, $\mathbf{T} = (0, \theta_t \cdot \mathbf{A}_\text{dir})$ and $\boldsymbol{q}_{r,t} = (1, \vec{0})$; for revolute joints, $\mathbf{T} = (0, \mathbf{A}_\text{pos})$ and $\boldsymbol{q}_{r,t} = \left(\cos(\theta_t/2), \sin(\theta_t/2) \cdot \mathbf{A}_\text{dir}\right)$. 
From the dual quaternion $\boldsymbol{q}_{r,t}$ and $\boldsymbol{q}_{d,t}$, the rotation matrix $\boldsymbol{R}_t$ and translation vector $\boldsymbol{t}_t$ can be derived for any timestep. The part mesh is deformed by $\mathbf{v}_m^{t} = \boldsymbol{R}_t \cdot \mathbf{v}_m + \boldsymbol{t}_t$, where $\mathbf{v}_m^t$ is the part mesh at timestep $t$.

To enable differentiable composition of the base and movable parts, we maintain their separate texture maps and render them independently, resulting in color images $\mathbf{I}_\text{base}$ and $\mathbf{I}_\text{mov}$, along with their corresponding depth maps $\mathbf{D}_\text{base}$ and $\mathbf{D}_\text{mov}$.
To resolve visibility conflicts between the rendered layers in a differentiable manner, we adopt \textit{Differentiable Soft Depth Blending}, computing per-pixel blending weights using a sigmoid function over depth difference.:
\begin{align}
\Delta \mathbf{D} &= \mathbf{D}_\text{mov} - \mathbf{D}_\text{base}, \\
\mathbf{w} &= \sigma(\Delta \mathbf{D} \cdot \beta), \\
\mathbf{I}_\text{pred} &= \mathbf{w} \cdot \mathbf{I}_\text{mov} + (1 - \mathbf{w}) \cdot \mathbf{I}_\text{base},
\end{align}
where $\sigma(\cdot)$ denotes the sigmoid function, and $\beta = 500.0$ is a sharpness parameter that controls the smoothness of the blending (higher values approximate hard z-buffering).
The resulting blended image $\mathbf{I}_\text{pred}$ serves as the final composited output. To supervise the optimization process, we minimize the discrepancy between $\mathbf{I}_\text{pred}$ and the reference image from the synthesized articulation video $V$. 
Finally, we use a visually complete part image (either base or movable), captured from the same view as the input view image, along with the amodal part mesh, for texture refinement~\cite{huang2024mv, zhao2025hunyuan3d}, aiming to enhance both texture realism and completeness.

\section{Experiments}
\subsection{Articulation Video Synthesis}
\label{sec:exp_video}
\begin{figure}[t]
  \includegraphics[width=\linewidth]{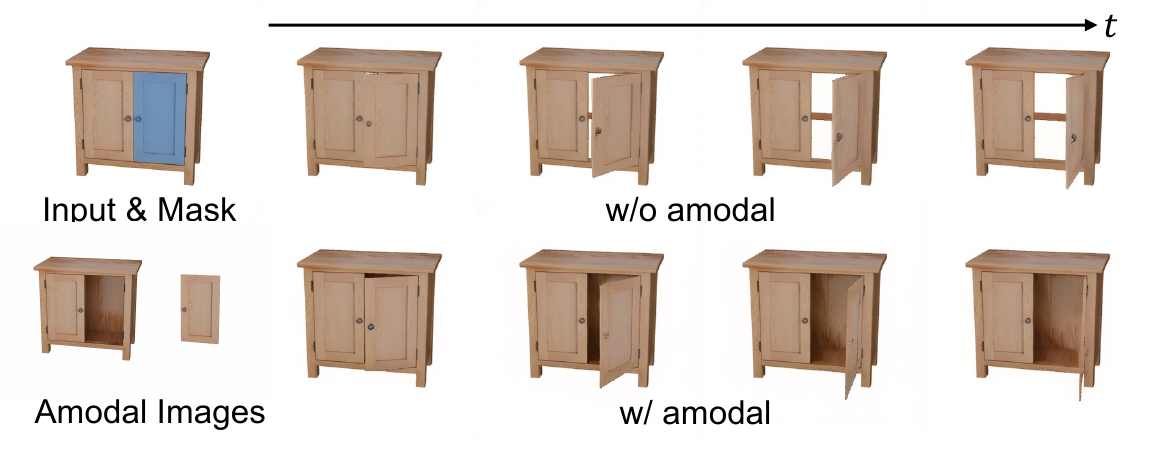}
  \caption{\textbf{Ablation on amodal images.} The inclusion of amodal images leads to more plausible articulation generation. }
  \label{fig:ablation}
\end{figure}

\paragraph{Setup} We construct the \underline{MA}sk-\underline{P}rom\underline{P}ted \underline{A}rticulation dataset (MAPPA) using articulated objects from PartNet-Mobility~\cite{xiang2020sapien, geng2023gapartnet, mo2019partnet}. For each video, we randomly sample the start and end configurations of an articulatable joint, applying heuristic constraints to ensure sufficient motion magnitude. In total, we generate 44k videos for training and 349 for testing. To enable photorealistic rendering, we leverage Phobos~\cite{von2020phobos} and Blender~\cite{blender}, utilizing the Cycles renderer with randomized environmental lighting. For fair evaluation, we adopt the same test split as DragAPart~\cite{li2024dragapart} and Puppet-Master~\cite{li2024puppet}, while re-rendering the sequences to obtain accurate movable part masks and more realistic visuals. For metrics, we utilize PSNR, SSIM~\cite{wang2004image}, and LPIPS~\cite{zhang2018unreasonable} to evaluate image alignment with ground-truth reference frames, and CLIP-T~\cite{esser2023structure} and FVD~\cite{unterthiner2018towards} to assess cross-frame consistency. 
\paragraph{Results} We demonstrate quantitative results in \cref{tab:video} and visual comparisons in \cref{fig:quali}, where our method outperforms prior baselines by a significant margin. We attribute this improvement to several key factors.
First, drag-based prompts often struggle to disambiguate the correct movable part in objects, especially in multi-part scenarios, leading to multiple parts moving simultaneously. Such issues are evident in the second example of both the in-domain and in-the-wild settings, where the drag prompt incorrectly influences multiple parts.
Second, the formulation of drag as a 2D point displacement in the input image introduces inherent ambiguities. Specifically, a single pixel in the image may correspond to multiple possible 3D locations, introducing undesirable biases for the model to disambiguate. For instance, in the first in-domain example, Puppet-Master incorrectly interprets the motion and attempts to merge the drawer with the upper structure.
Third, we find that the limited realism and fidelity of rendered training data used by Puppet-Master and DragAPart constrain its generalization capabilities to in-the-wild images. 
Additionally, we present an ablation study on the use of amodal images in \cref{fig:ablation}. Incorporating amodal guidance allows the model to focus solely on learning the articulation pattern, while occluded regions can be synthesized more accurately in accordance with the guidance. This will introduce minimal conflicts with the subsequent optimization process, integrating seamlessly into the overall asset generation pipeline.
\begin{table}[h]
\small
\centering
\caption{\textbf{Quantitative comparison on the test-split of MAPPA.}}
\resizebox{\linewidth}{!}{
\begin{tabular}{cccccc}
\toprule
Method & PSNR($\uparrow$) & SSIM($\uparrow$) & LPIPS($\downarrow$) & CLIP-T($\uparrow$) & FVD ($\downarrow$)\\
\midrule
DragAPart & 23.132 & 0.921 & 0.068 & \textbf{0.992} & 475.695 \\
Puppet-Master & 24.660 & 0.919 & 0.072 & 0.988 & 640.341 \\
Ours (w/o amodal) & 27.914 & 0.948 & 0.045 & 0.991 & 211.447 \\
Ours & \textbf{28.906} & \textbf{0.955} & \textbf{0.038} & \textbf{0.992} &  \textbf{195.707} \\
\bottomrule
\end{tabular}
}
\label{tab:video}
\end{table}

\subsection{Video-conditioned Asset Generation}
\label{sec:exp_asset}

\begin{figure}[t]
  \includegraphics[width=\linewidth]{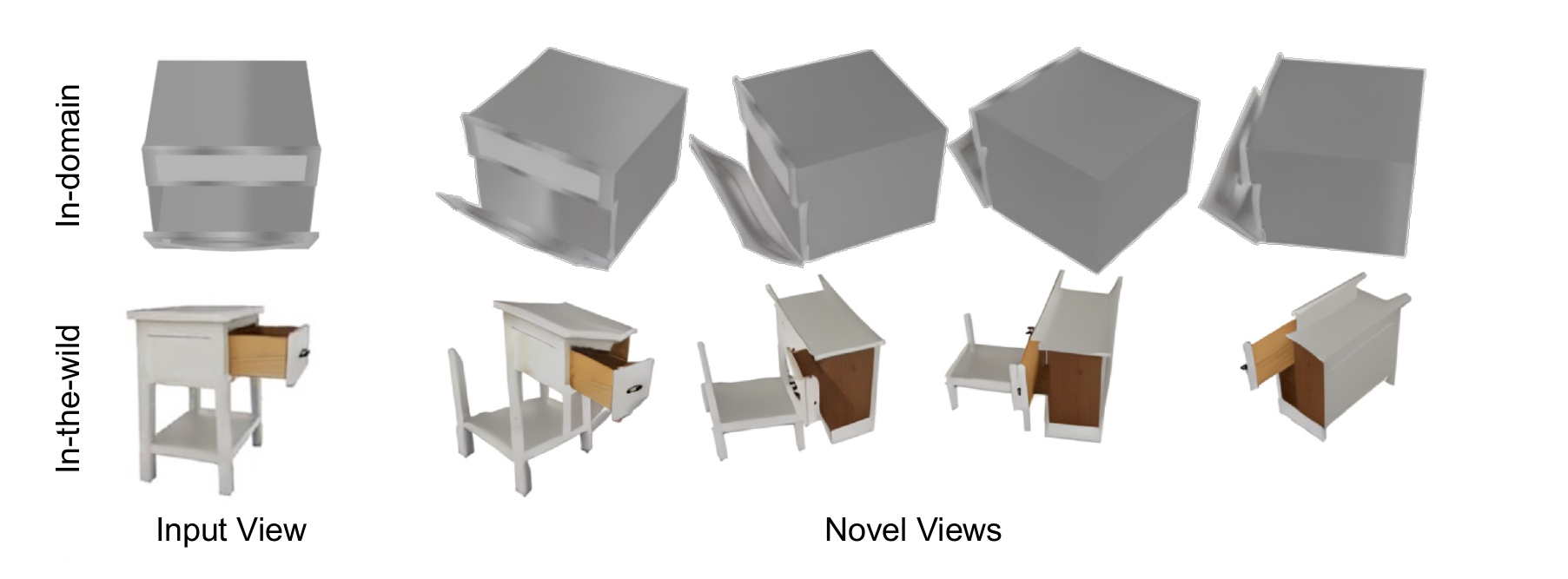}
  \caption{PartRM performs well on in-domain data but generalizes poorly to in-the-wild data. }
  \label{fig:partrm}
\end{figure}

\paragraph{Setup} We first apply a text-to-image generation model~\cite{ramesh2021zero} to obtain a reference image, and then synthesize a single-view articulation video using our proposed video generation model. We compare our method against two recent baselines in the Video-to-4D setting: STAG4D~\cite{zeng2024stag4d} and L4GM~\cite{ren2024l4gm}. We also consider PartRM~\cite{gao2025partrm}, which fine-tunes a large reconstruction model on PartNet-Mobility~\cite{xiang2020sapien}, but find that it generalizes poorly to in-the-wild data, as shown in \cref{fig:partrm} and therefore exclude it from our comparison. To assess the visual quality of the generated content without relying on reference images, we report Aesthetic Score (AS)~\cite{schuhmann2022laion} and MUSIQ~\cite{ke2021musiq}. In addition, we conduct a user study with 8 samples, where participants rate each sample on a 1–5 scale across three aspects: Articulation Plausibility (AP), which assesses whether the synthesized articulation is physically plausible; 3D Consistency (3DC), which evaluates whether the object appears coherent across multiple views; and Overall Quality (OQ), which reflects the perceived quality of the object in terms of geometry, appearance, and articulation. We collect a total of 304 responses from 38 participants.

\begin{table}[h]
\small
\centering
\caption{\textbf{Quantitative comparison on asset generation.}}
\begin{tabular}{cccccc}
\toprule
Method & AS($\uparrow$) & MUSIQ($\uparrow$) & AP($\uparrow$) & 3DC($\uparrow$) & OQ ($\uparrow$)\\
\midrule
L4GM & 4.473 & 57.745 & 3.039 & 2.706 & 2.627 \\
STAG4D & 4.119 & 32.678 & 2.768 & 2.465 & 2.333 \\
Ours & \textbf{4.632} & \textbf{70.968} & \textbf{4.641} & \textbf{4.612} & \textbf{4.602} \\
\bottomrule
\end{tabular}
\label{tab:asset}
\end{table}
\paragraph{Results} We present both quantitative results in \cref{tab:asset} and qualitative comparisons in \cref{fig:quali_comp}. Our method substantially outperforms baseline approaches in terms of image quality and user preference. We attribute this advantage to our usage of explicit mesh representation for the movable and base model, and our incorporation of articulation motion priors, which contribute to improved multi-view consistency and physically plausible articulated motion. While STAG4D and L4GM achieve reasonable alignment under the input video views, they exhibit noticeable degradation under novel views, with reduced quality in articulation dynamics, shape consistency, and appearance. This limitation stems from their reliance on off-the-shelf multi-view image generators and the lack of explicit modeling of articulation motion in their pipelines.

\section{Conclusion}
\paragraph{Limitation} Our method builds on off-the-shelf image-to-3D generation models, which may occasionally produce physically implausible results, such as cabinet doors that are unrealistically short or long when open.
Also, articulation optimization from a single-view video is sensitive to viewpoint ambiguities and occlusions, which we plan to address with multi-view video generation in future work.

In summary, we propose a novel framework for generating articulated objects that combines part-aware 3D object generation, single-view articulation video synthesis, and articulation joint estimation. Our method produces accurately segmented parts, high-quality geometry and texture, and physically plausible articulation behaviors, facilitating downstream tasks like editing and robotics.

\bibliographystyle{ACM-Reference-Format}
\bibliography{main}
\begin{figure*}[h]
  \includegraphics[width=\linewidth]{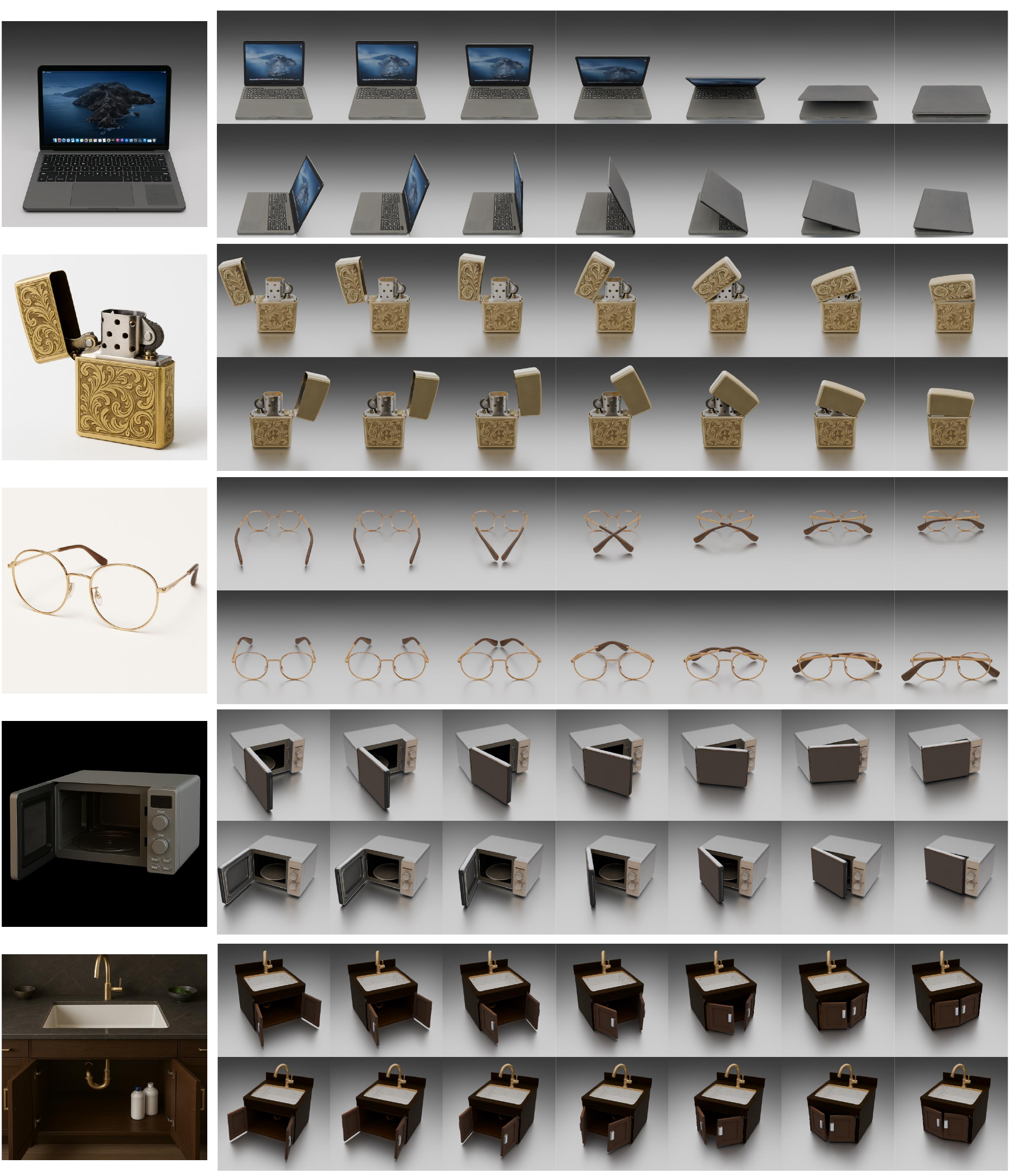}
  \caption{\textbf{Multi-view renderings of articulations.}}
  \label{fig:pure_img_0}
\end{figure*}
\begin{figure*}[h]
  \includegraphics[width=\linewidth]{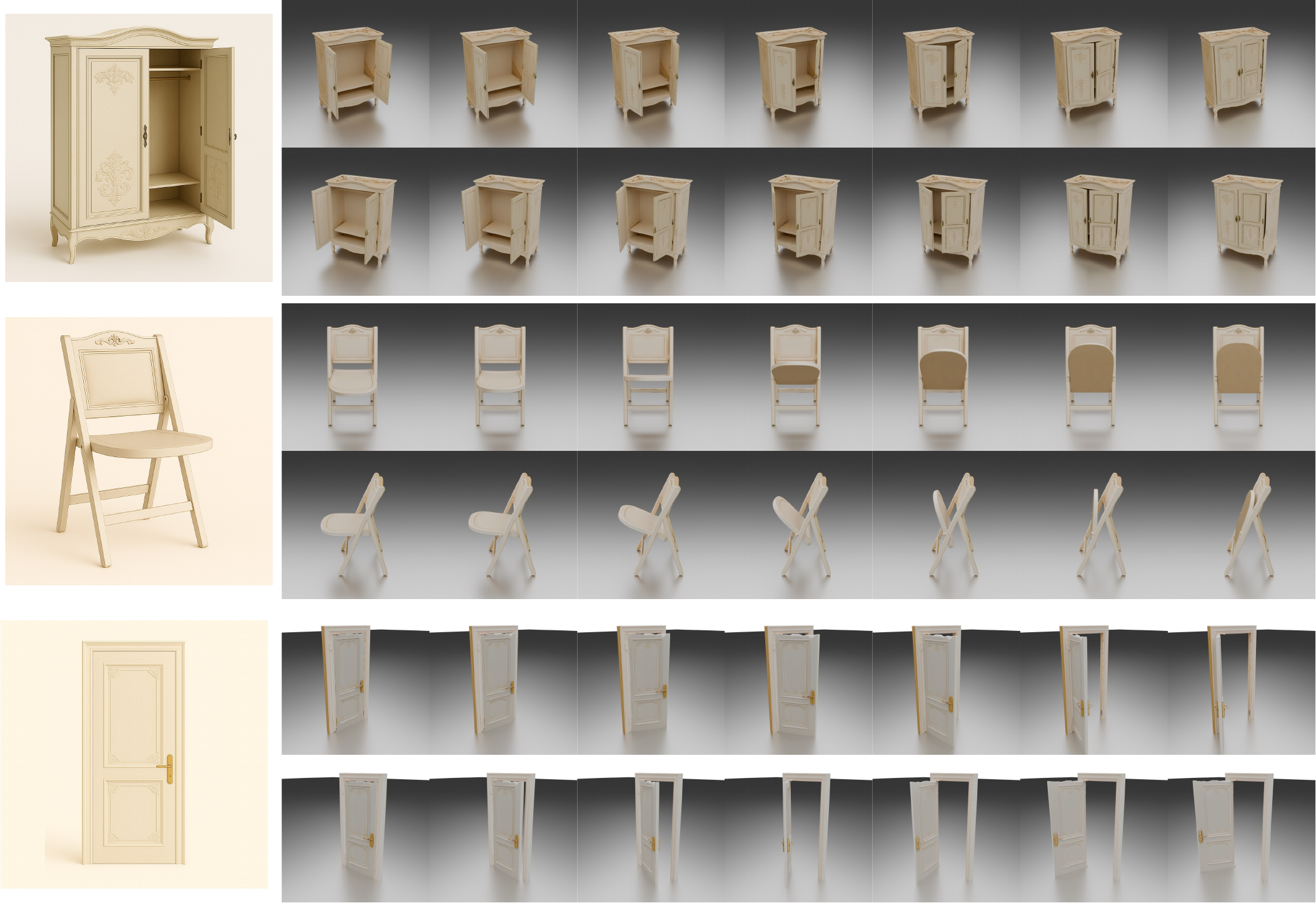}
  \caption{\textbf{Multi-view renderings of articulations.}}
  \label{fig:pure_img_1}
\end{figure*}
\begin{figure*}[h]
  \includegraphics[width=\linewidth]{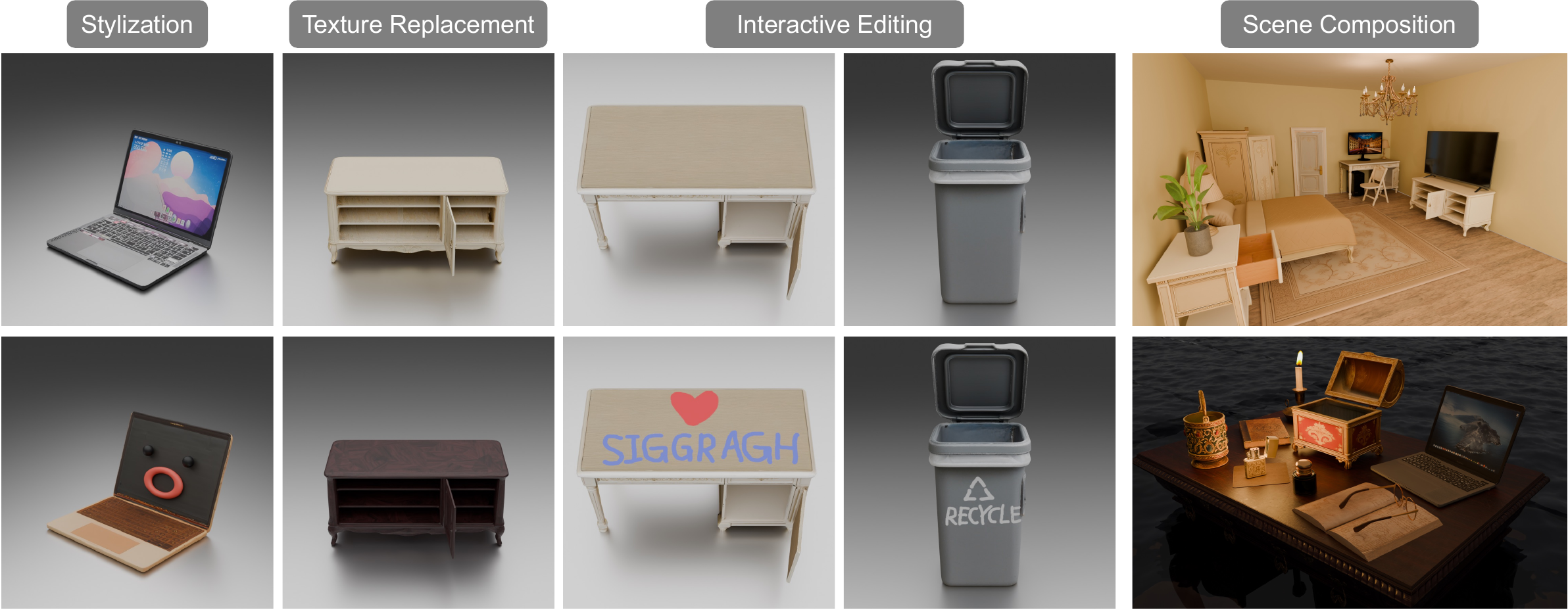}
  \caption{\textbf{Downstream Tasks.} Our framework supports asset editing through texture and geometry modifications, and enables scene composition using the generated articulated objects.}
  \label{fig:pure_img_2}
\end{figure*}

\end{document}